\documentclass[11pt,a4paper]{article}
\usepackage{times,latexsym}
\usepackage{url}

\usepackage[table,xcdraw]{xcolor}
\usepackage{graphicx}

\usepackage[acceptedWithA]{tacl2021v1}

\usepackage{xspace,mfirstuc,tabulary}

\newif\iftaclinstructions
\taclinstructionsfalse 
\iftaclinstructions
\renewcommand{\confidential}{}
\renewcommand{\anonsubtext}{(No author info supplied here, for consistency with
TACL-submission anonymization requirements)}
\newcommand{\instr}
\fi

\iftaclpubformat 

\else

\fi

\usepackage{soul}

\usepackage{booktabs}

\newcommand{\ctext}[3][RGB]{
  \begingroup
  \definecolor{hlcolor}{#1}{#2}\sethlcolor{hlcolor}
  \hl{#3}%
  \endgroup
}

\usepackage[utf8]{inputenc}

\usepackage{adjustbox}
\usepackage{microtype}
\usepackage{listings}
\usepackage{multirow}
\usepackage{multicol}
\usepackage{makecell}

\usepackage{graphicx}
\usepackage{linguex}
\usepackage{subfigure}
\usepackage{amsmath}

\usepackage{hyperref}
\usepackage{enumitem}

\usepackage{amssymb}
\usepackage{pifont}

\RequirePackage{etoolbox}
\RequirePackage{xcolor}
\definecolor{Gray}{gray}{0.9}
\definecolor{cb-blue-green} {RGB}{ 0,  073,  073}
\definecolor{cb-green-sea}  {RGB}{ 0, 146, 146}
\definecolor{cb-rose}       {RGB}{255, 109, 182}
\definecolor{cb-salmon-pink}{RGB}{255, 182, 119}
\definecolor{cb-purple}     {RGB}{ 73,   0, 146}
\definecolor{cb-blue}       {RGB}{ 0, 109, 219}
\definecolor{cb-lilac}      {RGB}{182, 109, 255}
\definecolor{cb-blue-sky}   {RGB}{109, 182, 255}
\definecolor{cb-blue-light} {RGB}{182, 219, 255}
\definecolor{cb-burgundy}   {RGB}{146,   0,   0}
\definecolor{cb-brown}      {RGB}{146,  73,   0}
\definecolor{cb-clay}       {RGB}{219, 209,   0}
\definecolor{cb-green-lime} {RGB}{ 36, 255,  36}
\definecolor{cb-yellow}     {RGB}{255, 255, 109}

\newcommand{\xmark}{\ding{55}}

\title{mGPT: Few-Shot Learners Go Multilingual}

\author{
Oleh Shliazhko$^1$\thanks{\ \ Work done while at SaluteDevices.}, Alena Fenogenova$^2$, Maria Tikhonova$^{2,3}$, \\ \textbf{Anastasia Kozlova}$^2$, \textbf{Vladislav Mikhailov}$^{2*}$\thanks{\ \ Now at University of Oslo.}, \textbf{Tatiana Shavrina}$^{2,4,5,6}$$^*$
   \\
   $^1$Independent Researcher, $^2$SaluteDevices, Russia, $^3$HSE University, Russia, $^4$AIRI, Russia
   \\
   $^5$AI Center, NUST MISiS, Russia, $^6$Institute of Linguistics RAS, Russia \\
    \texttt{olehshliazhko@gmail.com, alenush93@gmail.com, 
 mtihonova@hse.ru,} \\
   \texttt{anastasi2510@gmail.com, vvmkhlvv@gmail.com, rybolos@gmail.com}
}

\date{}

\begin{document}
\maketitle
\begin{abstract}
  This paper introduces mGPT, a multilingual variant of GPT-3, pretrained on $61$ languages from linguistically diverse $25$ language families using Wikipedia and C4 Corpus. We detail the design and pretraining procedure. The models undergo an intrinsic and extrinsic evaluation: language modeling in all languages, downstream evaluation on cross-lingual NLU datasets and benchmarks in 33 languages, and world knowledge probing in 23 languages. The in-context learning abilities are on par with the contemporaneous language models while covering a larger amount of languages, including underrepresented and low-resource languages of the Commonwealth of Independent States and the small peoples in Russia. The source code and the language models are publicly available under the MIT license.
\end{abstract}

\section{Introduction}

The advent of the Transformer architecture~\cite{vaswani2017attention} has facilitated the development of various language models (LMs;~\citealp{liu2020survey}). Although the well-established ``pretrain \& finetune'' paradigm has led to rapid progress in NLP~\cite{wang2019superglue}, it imposes several limitations. Finetuning relies on an extensive amount of labeled data. Collecting high-quality labeled data for new tasks and languages is expensive and resource-consuming~\cite{wang-etal-2021-want-reduce}. LMs can learn spurious correlations from finetuning data~\cite{naik-etal-2018-stress,niven-kao-2019-probing} and demonstrate inconsistent generalization, catastrophic forgetting, or brittleness to finetuning data order~\cite{mccoy-etal-2020-berts,dodge2020finetuning}. Last but not least, finetuning requires additional computational resources and, therefore, aggravates the problem of a large carbon footprint~\cite{bender2021dangers}.

The latest approaches address these limitations with zero-shot and few-shot learning, performing a task with LM scoring or conditioning on a few demonstration examples without parameter updates~\cite{NEURIPS2020_1457c0d6}. Autoregressive LMs adopted via these paradigms have been widely applied in many NLP tasks~\cite{schick-schutze-2021-just,perez2021true}, notably in cross-lingual knowledge transfer~\cite{winata-etal-2021-language} and low-resource language scenarios~\cite{lin2022fewshot}. However, model development for underrepresented typologically distant and low-resource languages~\cite{wu-dredze-2020-languages,lauscher-etal-2020-zero,hedderich-etal-2021-survey} and cross-lingual generalization abilities of autoregressive LMs~\cite{erdem2022neural} have been left understudied.

This paper presents mGPT, a multilingual version of GPT-3~\cite{NEURIPS2020_1457c0d6} available in $1.3$B (mGPT$_\text{1.3B}$) and $13$B (mGPT$_\text{13B}$) parameters. We aim to (i) develop a large-scale multilingual autoregressive LM that inherits the GPT-3's generalization benefits and (ii) to increase the linguistic diversity of multilingual LMs, making the first attempt to address languages of the Commonwealth of Independent States (CIS) and under-resourced languages of the small peoples in Russia. We pretrain mGPT in $61$ languages from $25$ language families on Wikipedia and Colossal Clean Crawled Corpus (C4;~\citealp{raffel2020exploring}). We analyze the mGPT's performance on various intrinsic and extrinsic tasks and compare it with the contemporaneous generative LMs. 

\paragraph{Key findings} The analysis reveals that (i) mGPT$_\text{1.3B}$ is comparable to XGLM$_\text{1.7B}$~\cite{lin2022fewshot} while having fewer weights and covering a larger amount of languages, (ii) mGPT shows confident performance on Austronesian, Austro-Asiatic, Japonic, Germanic, and Romance languages on multiple tasks and prominent language modeling abilities on the languages of the small peoples in Russia, (iii)
adding more demonstrations may result in performance degradation for both mGPT and XGLM, and (iv) hate speech detection is one of the most challenging tasks, receiving random guessing performance in the zero-shot and few-shot evaluation setups. External validation by the NLP community since the release\footnote{As of the time of writing this paper, mGPT$_\text{1.3B}$ was publicly available. Note that mGPT$_\text{13B}$ is also now released.} shows that mGPT$_\text{1.3B}$ can outperform large-scale LMs on SuperGLUE tasks and promote strong solutions for multilingual clause-level morphology tasks. We release the model evaluation code\footnote{\href{https://github.com/ai-forever/mgpt}{\texttt{github.com/ai-forever/mgpt}}}, the mGPT$_\text{1.3B}$\footnote{\href{https://huggingface.co/ai-forever/mGPT}{\texttt{hf.co/ai-forever/mGPT}}} and mGPT$_\text{13B}$\footnote{\href{https://huggingface.co/ai-forever/mGPT-13B}{\texttt{hf.co/ai-forever/mGPT-13B}}} models. We hope to facilitate research on the applicability of autoregressive LMs in non-English languages and increase the linguistic inclusivity of the low-resource languages.

\begin{figure*}[!ht]
    \centering    \includegraphics[width=.95\textwidth]{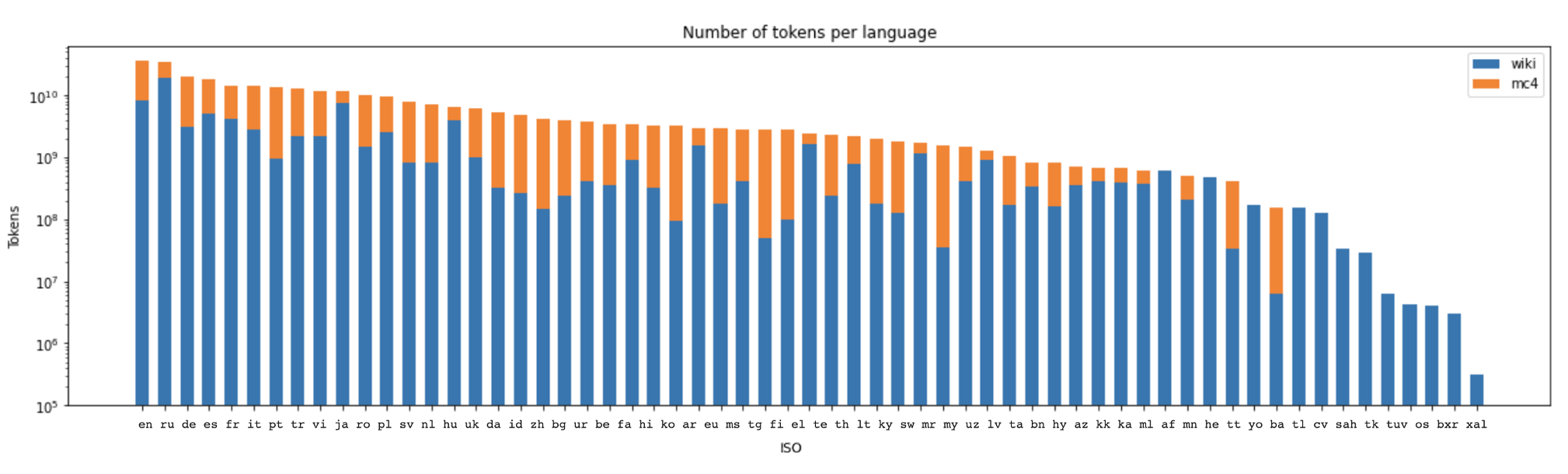}
    \caption{Number of tokens for each language in the pretraining corpus on a logarithmic scale.}
    \label{fig:dataset_stats}
\end{figure*}

\begin{figure*}[t!]
  \centering \includegraphics[width=.95\textwidth]{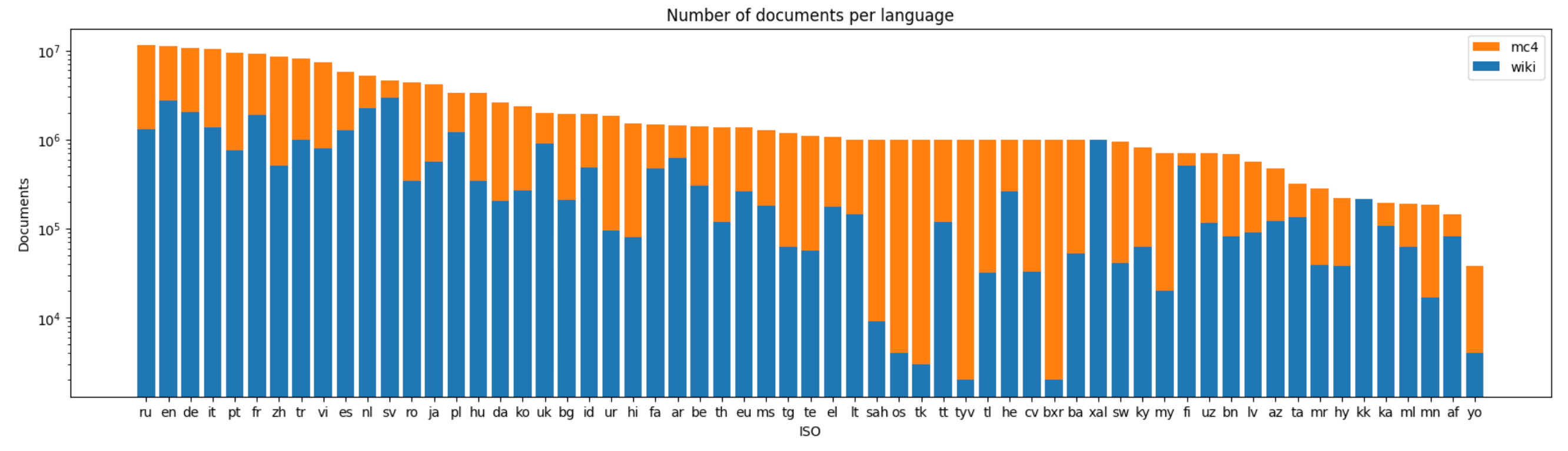}
  \caption{Number of documents for each language in the pretraining corpus on a logarithmic scale.}
  \label{fig:doc_number}
\end{figure*}

\section{Related Work}
\label{sec:previous_work}
\paragraph{Multilingual Transformers} Recent years have featured the development of various monolingual and multilingual LMs initially designed for English. BERT~\cite{devlin-etal-2019-bert} has been replicated in other high-resource languages~\cite{martin-etal-2020-camembert,masala-etal-2020-robert} and language families, e.g., Indian~\cite{kakwani-etal-2020-indicnlpsuite} and Balto-Slavic~\cite{arkhipov-etal-2019-tuning}. Massively multilingual LMs -- mBERT, XLM-R~\cite{conneau-etal-2020-unsupervised}, RemBERT~\cite{chung2021rethinking}, mBART~\cite{liu-etal-2020-multilingual-denoising} and mT5~\cite{xue-etal-2021-mt5} -- have now pushed state-of-the-art results on various NLP tasks in multiple languages~\cite{kalyan2021ammus}. Such models support more than $100$ languages and vary in the architecture design and pretraining objectives. By contrast, our work presents one of the first multilingual \emph{autoregressive} LMs covering more than 61 languages.

\paragraph{GPT-based Language Models} Large-scale generative LMs (e.g., GPT-3;~\citealp{NEURIPS2020_1457c0d6}) are triggering a shift from the ``pretrain \& finetune'' paradigm to prompt-based learning~\cite{liu2023pre}. The benefit of balancing the pretraining costs and performing standardized NLP tasks with a few demonstration examples has stimulated the development of open-source autoregressive LMs for English~\citep[e.g.,][]{black-etal-2022-gpt,biderman2023pythia,dey2023cerebrasgpt}, Chinese~\cite{zeng2021pangualpha}, and Russian~\cite{zmitrovich2023family}. A few contemporaneous works extend the research on zero-shot and few-shot learning, evaluating the in-context abilities of GPT-based LMs in multilingual scenarios.~\citet{winata-etal-2021-language} report that English GPTs perform significantly better than random guessing with monolingual and multilingual prompts on typologically close languages, such as French, Spanish, and German. ~\citet{lin2022fewshot} propose XGLM, a multilingual GPT-style LM in $30$ languages, and empirically show that it can outperform its monolingual counterparts of the comparable number of parameters. We use XGLM as the main baseline in our experiments and analyze the results of comparing mGPT$_\text{1.3B}$ with other autoregressive LMs published after our release, such as BLOOM~\cite{workshop2023bloom}.

\begin{table}[t!]
\centering
\tiny
\begin{tabular}{ll}
\toprule
\textbf{Language Family}  & \textbf{Languages} \\
\midrule
Afro-Asiatic &  Arabic (ar), Hebrew (he)\\
Austro-Asiatic &    Vietnamese (vi)\\
\multirow{2}{*}{Austronesian} &  Indonesian (id), Javanese (jv), Malay (ms)   \\ & Tagalog (tl) \\
Baltic & Latvian (lv), Lithuanian (lt) \\
Basque & Basque (eu) \\
Dravidian & Malayalam (ml), Tamil (ta), Telugu (te) \\
Indo-European (Armenian) & Armenian (hy) \\
\multirow{2}{*}{Indo-European (Indo-Aryan)} & Bengali (bn), Marathi (mr), Hindi (hi), \\ & Urdu (ur) \\
\multirow{2}{*}{Indo-European (Germanic)} & Afrikaans (af), Danish (da), English (en), \\ & German (de), Swedish (sv) \\
\multirow{2}{*}{Indo-European (Romance)} & French (fr), Italian (it), Portuguese (pt), \\ & Romanian (ro), Spanish (es) \\
Indo-European (Greek) &  Greek (el) \\
Indo-European (Iranian) & Ossetian (os), Tajik (tg), Persian (fa) \\
Japonic &  Japanese (ja) \\
Kartvelian &  Georgian (ka) \\
Koreanic &  Korean (ko) \\
Kra-Dai & Thai (th) \\
Mongolic &  Buryat (bxr), Kalmyk (xal), Mongolian (mn) \\
Niger-Congo &     Swahili (sw), Yoruba (yo) \\
\multirow{2}{*}{Slavic} &  Belarusian (be), Bulgarian (bg), Russian (ru), \\ & Ukrainian (uk), Polish (pl) \\
Sino-Tibetan &   Burmese (my)\\ 
Turkic (Karluk) &  Uzbek (uz)\\
\multirow{2}{*}{Turkic (Kipchak)} &   Bashkir (ba), Kazakh (kk), Kyrgyz (ky), \\ & Tatar (tt) \\
\multirow{2}{*}{Turkic (Oghuz)} &     Azerbaijani (az), Chuvash (cv), Turkish (tr), \\ & Turkmen (tk) \\
Turkic (Siberian) &   Tuvan (tyv), Yakut (sax) \\
Uralic &  Estonian (et), Finnish (fi), Hungarian (hu) \\
\bottomrule
\end{tabular}
\caption{A list of languages by the language family.}
\label{tab:languages}
\end{table}

\section{Method}
\label{sec:method}
\subsection{Pretraining Data}
\label{sec:data}
\paragraph{Language Selection} ~\autoref{tab:languages} summarizes the list of languages by their family. The pretraining corpus consists of a typologically weighted set of languages covered by cross-lingual benchmarks, such as XGLUE~\cite{liang-etal-2020-xglue} and XTREME~\cite{hu2020xtreme}. The motivation behind the language choices is to narrow the gap between the high-resource and low-resource languages~\cite{ducel-etal-2022-name}. To this end, we include 20 languages from the tail of the C4 language list, the list of underrepresented languages of Russia, and the official and resource-lean CIS languages~\cite{orekhov2016russian}.

\paragraph{Data Preparation Pipeline} Pretraining extensive LMs requires large volumes of high-quality data. Despite the explosive growth of web corpora resulting in the pretraining data volume of up to 6T tokens~\cite{xue-etal-2021-mt5}, the data quality is often unsatisfactory~\cite{kreutzer-etal-2022-quality}. General approaches to maximizing the quality are based on manually curated heuristics~\cite{yang2019xlnet}, the perplexity of LMs~\cite{wenzek-etal-2020-ccnet}, and data quality classifiers~\cite{NEURIPS2020_1457c0d6}. Our data preparation pipeline includes data collection,  deduplication, and filtration.

\paragraph{Data Collection} The pretraining corpus represents a collection of documents from Wikipedia and C4. The Wikipedia texts are extracted from the dumps (v. $20201101$) with WikiExtractor~\cite{Wikiextractor2015}. The C4 data is downloaded using the Tensorflow datasets\footnote{\href{https://tensorflow.org/datasets/catalog/c4}{\texttt{tensorflow.org/datasets/catalog/c4}}}~\cite{paper2021tensorflow}.

\paragraph{Deduplication} The text deduplication includes 64-bit hashing of each text in the pretraining corpus for keeping texts with a unique hash.

\paragraph{Filtration} We follow~\citet{OrtizSuarezSagotRomary2019} on the C4 data filtration. We also filter the documents based on their text compression rate using \texttt{zlib}\footnote{\href{https://docs.python.org/3/library/zlib.html}{\texttt{docs.python.org/3/library/zlib}}}. The most strongly and weakly compressing deduplicated texts are discarded. The compression range for an acceptable text is empirically defined as ×$1.2$ — ×$8$. The texts with an entropy of less than $1.2$ contain code junk and entities, while those of more than $8$ contain repetitive segments. The next step includes distinguishing between low and high-quality documents with a binary classifier. The classifier is trained with Vowpal Wabbit\footnote{\href{https://github.com/VowpalWabbit/vowpal_wabbit}{\texttt{github.com/VowpalWabbit/vowpal\_wabbit}}} on the Wikipedia documents as positive examples and the filtered C4 documents as negative ones. The remainder is cleaned by a set of language-agnostic heuristics. The size of the pretraining corpus is $46$B (Wikipedia), and $442$B UTF characters (C4), resulting in 600GB. ~\autoref{fig:dataset_stats} shows the total number of tokens for each language, and the total number of documents in the pretraining corpus is presented in~\autoref{fig:doc_number}.

\subsection{Tokenization}
\label{section:tokenization}
The design of the tokenization method may have a significant impact on learning efficient representations, model memorization, and downstream performance~\cite{mielke2021words,nogueira2021investigating,pfeiffer-etal-2021-unks,rust-etal-2021-good}. We investigate the effect of the tokenization strategy on the model perplexity. We pretrain five strategy-specific versions of mGPT$_\text{163M}$ on a Wikipedia subset of the pretraining corpus. The tokenization strategy is selected based on their perplexity on a held-out Wikipedia sample (approx. $10.7$MB), which is inferred as~\autoref{eqn:ppl}.

\vspace{-15pt}
\begin{equation}
  PPL(t) = exp(-\frac{1}{|c|}\sum_{i=0}^{|t|} 
log_{p_{\theta}}(x_i|x_{<i}))
\label{eqn:ppl}
\end{equation}

where $t$ is an input text, $|t|$ is the length of the text in tokens, $|c|$ is the length of the text in characters. The perplexity is normalized over the number of characters since the tokenizers produce different numbers of tokens for $t$~\cite{cotterell-etal-2018-languages}. 

\begin{table}[t!]
    \centering
    \scriptsize
    \begin{tabular}{ll}
         \toprule
         \textbf{Strategy} & \textbf{Tokenization Example} \\ \midrule
         \textsc{default} & \ctext[RGB]{219, 209,   0}{22}, \ctext[RGB]{219, 209,   0}{ Birds}, \ctext[RGB]{219, 209,   0}{ +}, \ctext[RGB]{219, 209,   0}{ 3}, \ctext[RGB]{219, 209,   0}{ birds}, \ctext[RGB]{219, 209,   0}{ =}, \ctext[RGB]{219, 209,   0}{ 25}, \ctext[RGB]{219, 209,   0}{ birds} \\
         \textsc{case} & \ctext[RGB]{255, 255, 109}{22},
          \ctext[RGB]{255, 255, 109}{\texttt{<case>}}, \ctext[RGB]{255, 255, 109}{birds},  \ctext[RGB]{255, 255, 109}{ +},  \ctext[RGB]{255, 255, 109}{ 3}, \ctext[RGB]{255, 255, 109}{ birds}, $...$ \\

         \textsc{arithmetic} & \ctext[RGB]{0, 146, 146}{2}, \ctext[RGB]{0, 146, 146}{2}, \ctext[RGB]{0, 146, 146}{\texttt{<case>}}, \ctext[RGB]{0, 146, 146}{birds}, \ctext[RGB]{0, 146, 146}{ }, \ctext[RGB]{0, 146, 146}{+}, \ctext[RGB]{0, 146, 146}{ }, \ctext[RGB]{0, 146, 146}{3},  $...$ \\

         \textsc{combined} & \ctext[RGB]{109, 182, 255}{2}, \ctext[RGB]{109, 182, 255}{2}, \ctext[RGB]{109, 182, 255}{\texttt{<case>}}, \ctext[RGB]{109, 182, 255}{birds}, \ctext[RGB]{109, 182, 255}{ }, \ctext[RGB]{109, 182, 255}{+}, \ctext[RGB]{109, 182, 255}{ }, \ctext[RGB]{109, 182, 255}{3}, \ctext[RGB]{109, 182, 255}{ }, $...$ \\
         
         \textsc{char} & \ctext[RGB]{182, 219, 255}{2}, \ctext[RGB]{182, 219, 255}{2}, \ctext[RGB]{182, 219, 255}{ }, \ctext[RGB]{182, 219, 255}{B}, \ctext[RGB]{182, 219, 255}{i}, \ctext[RGB]{182, 219, 255}{r}, \ctext[RGB]{182, 219, 255}{d}, \ctext[RGB]{182, 219, 255}{s}, \ctext[RGB]{182, 219, 255}{ }, \ctext[RGB]{182, 219, 255}{+}, \ctext[RGB]{182, 219, 255}{ }, $...$
         \\ \bottomrule
    \end{tabular}
    \caption{Different tokenization strategies applied to the sentence ``22 Birds + 3 birds = 25 birds''. The resulting tokens are highlighted in the corresponding colors.}
    \label{table:tokenizer_strategies}
\end{table}

\paragraph{Tokenization Strategies}
We considered five tokenization strategies incorporating specific representations of uppercase characters, numbers, punctuation marks, and whitespaces.~\autoref{table:tokenizer_strategies} presents examples of the tokenization strategies. 

\begin{itemize}[itemsep=-2pt,partopsep=1ex,parsep=1ex,leftmargin=1.5em]
    \item \textsc{default}: BBPE~\cite{wang2020neural};
    \item \textsc{case}: Each uppercase character is replaced with a special token \texttt{<case>} followed by the corresponding lowercase character;
    \item \textsc{arithmetic}: The \textsc{case} strategy combined with representing numbers and arithmetic operations as individual tokens;
    \item \textsc{combined}: The \textsc{arithmetic} strategy combined with representing punctuation marks and whitespaces as individual tokens;
    \item \textsc{char}: Character-level tokenization.
\end{itemize}

\paragraph{Pretraining Details} The models are pretrained on $16$ V100 GPUs for $600$k training steps with a set of fixed hyperparameters: vocabulary size of $100$k, context window of $2048$, learning rate of $2\emph{e}^{-4}$, and batch size of $4$.

\paragraph{Results} The experiment results are presented in~\autoref{table:eval_ppl}. The \textsc{default} model achieves the best results, outperforming the rest of the models by up to $2.5$ of perplexity score. Based on this experiment, we select the \textsc{default} strategy to pretrain the mGPT$_\text{1.3B}$ and mGPT$_\text{13B}$ models.

\begin{table}[t!]
    \centering
    \small
    \begin{tabular}{lc}
    \toprule
    \textbf{Strategy} & \textbf{Avg. PPL}
    \\\midrule
    \textsc{default} & \textbf{6.94} \\
    \textsc{case} & 8.13 \\
    \textsc{arithmetic} & \underline{7.99} \\
    \textsc{combined} & 8.43 \\
    \textsc{char} & 9.47 \\ \bottomrule
    \end{tabular}
    \caption{The average perplexity results. The best score is put in bold, the second best is underlined.}
    \label{table:eval_ppl}
\end{table}

\begin{table}[t!]
\centering
\small
\begin{tabular}{lccc}
\toprule
\textbf{Model}  & \textbf{Size} & \textbf{Layers} & $d_{model}$ \\ \midrule
GPT-2      & 1.5B  & 48      & 1600   \\ 
GPT-3$_\text{1.3B}$   & 1.3B    & 24      & 2048   \\ 
GPT-3$_\text{13B}$   & 13B    & 40      & 5120   \\ \bottomrule
\end{tabular}
\caption{Comparison of GPT-2 and GPT-3. The mGPT architecture replicates the parameters of GPT-3$_\text{1.3B}$ and GPT-3$_\text{13B}$, and uses sparse attention in alternating dense and sparse layers.}
\label{table:models}
\end{table}
\subsection{Model Architecture}
The mGPT architecture is based on GPT-3. We use the architecture description by~\citeauthor{NEURIPS2020_1457c0d6}, the GPT-2 code base~\cite{radford2019language} from HuggingFace~\cite{wolf-etal-2020-transformers} and Megatron-LM~\cite{shoeybi2020megatronlm}.~\autoref{table:models} presents the description of the GPT-2 and GPT-3 architectures of comparable sizes. With all the other hyperparameters equal, GPT-3 has fewer layers (\textit{Layers}: $48$ vs. $24$) but a larger hidden size ($d_{model}$: $1600$ vs. $2048$) as opposed to GPT-2. GPT-3 also alternates the classic dense and sparse attention layers~\cite{child2019generating}. 

\begin{figure*}[!ht]
  \centering
  \includegraphics[width=.9\textwidth]{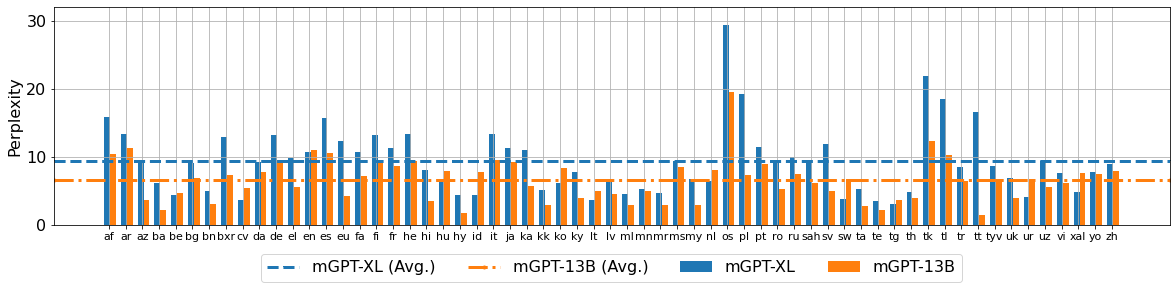}
  \caption{Language-wise perplexity results. Lower is better.}
  \label{fig:lang_perplexities}
\end{figure*}

\begin{figure*}[t!]
  \centering
  \includegraphics[width=.9\textwidth]{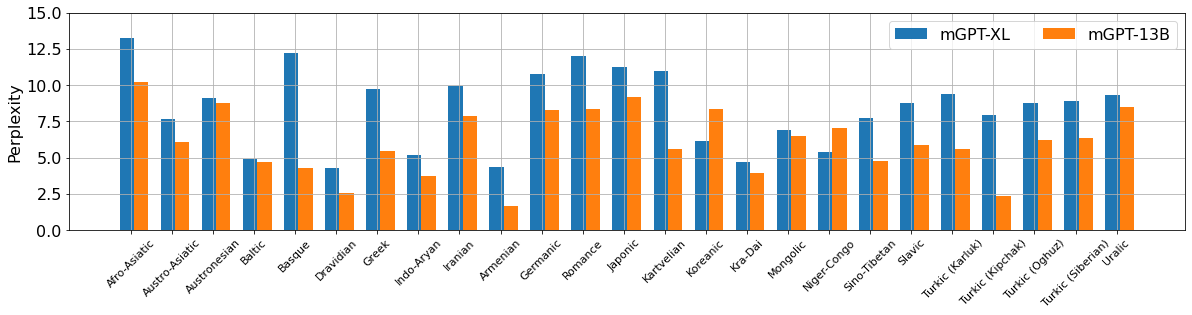}
  \caption{Family-wise perplexity results. The scores are averaged over the number of languages within each family.}
  \label{fig:family_wise_ppl}
\end{figure*}

\subsection{Model Pretraining}
The pretraining procedure mostly follows~\citeauthor{NEURIPS2020_1457c0d6}. 
We utilize the DeepSpeed library~\cite{rasley2020deepspeed} and Megatron-LM~\cite{shoeybi2020megatronlm}. We pretrain our LMs with a total batch size of $2048$ and a context window of $512$ tokens. The total number of the training steps is $600$k, and the models have seen $400$B tokens during pretraining. The pretraining took $14$ days on a cluster of $256$ V100 GPUs for mGPT$_\text{1.3B}$ and $22$ days on $512$ V100 GPUs for mGPT$_\text{13B}$. We report the computational, energy, and carbon costs in~\S\ref{ethical:co2}.

\section{Experiments}
\subsection{Language Modeling}
\label{subsection:language_modeling}
\paragraph{Method} We estimate the language modeling performance on the held-out sets for each language. Here, perplexity is computed as described in \S\ref{section:tokenization}, except that perplexity is normalized over the length of the input text $t$ in tokens $|t|$. We also run statistical tests to analyze the effect of linguistic, dataset, and model configuration criteria:

\begin{itemize}[itemsep=-2pt,partopsep=1ex,parsep=1ex,leftmargin=1.5em]
    \item \textit{Language script}: we divide the languages into two groups by their script -- Latin and others (e.g., Cyrillic and Arabic) -- and use the Mann-Whitney U test~\cite{mann1947controlling} to analyze the perplexity distributions in the groups.
    \item \textit{Pretraining corpus size}: we calculate the Pearson correlation coefficient~\cite{pearson1895vii} to analyze the correlation between the language perplexity and the number of documents in this language in the pretraining corpus.
    \item \textit{Model size}: we use the Mann-Whitney U test to analyze the effect of the model size.
\end{itemize}

\paragraph{Results by Language} ~\autoref{fig:lang_perplexities} presents the perplexity scores for each language on the held-out sets. The mGPT$_\text{13B}$ model achieves the best perplexities within the 2-to-10 score range for the majority of languages, including Dravidian (Malayalam, Tamil, Telugu), Indo-Aryan (Bengali, Hindi, Marathi), Slavic (Belarusian, Ukrainian, Russian, Bulgarian), Sino-Tibetan (Burmese), Kipchak (Bashkir, Kazakh) and others. Higher perplexities up to 20 are for only seven languages from different families. The mGPT$_\text{1.3B}$ results have similar distribution but are consistently higher than mGPT$_\text{13B}$. 

\begin{table}[t!]
\centering
\scriptsize
\begin{tabular}{lccc}
\toprule
\textbf{Criterion}  & \textbf{Model} & \textbf{Test} & p-value \\ \midrule
Language script      & \begin{tabular}{@{}c@{}}mGPT$_\text{1.3B}$ \\ mGPT$_\text{13B}$\end{tabular}  & M-W U test & \begin{tabular}{@{}c@{}} 0.012 \\ 0.000 \end{tabular} \\ 

Pretraining corpus size      & \begin{tabular}{@{}c@{}}mGPT$_\text{1.3B}$ \\ mGPT$_\text{13B}$\end{tabular}  & Pearson & \begin{tabular}{@{}c@{}} 0.137 \\ 0.307 \end{tabular} \\ 

Model size     & \begin{tabular}{@{}c@{}}mGPT$_\text{1.3B}$ \\ mGPT$_\text{13B}$\end{tabular}  & M-W U test & 0.0007 \\ 

\bottomrule
\end{tabular}
\caption{Correlation analysis results.}
\label{table:corr_res}
\end{table}
\begin{table*}[th!]
    \centering
    \scriptsize
    \resizebox{\textwidth}{!}{
    \begin{tabular}{llr}
    \toprule
    \textbf{Task} & \textbf{Template} & \textbf{Output Candidates}
    \\ \midrule

    \multirow{2}{*}{\textbf{XNLI}} & \multirow{2}{*}{
    \texttt{<s> \{sentence 1\}}, right? \texttt{\{label\} \{sentence 2\} </s>}} & 
    
    Yes (Entailment); Also (Neutral) \\ & & No (Contradiction) \\ \midrule
    
    \textbf{PAWSX} &
    \texttt{<s> \{sentence 1\}}, right? \texttt{\{label\} \{sentence 2\} </s>} & Yes; No \\ \midrule
    
    \textbf{XWINO} &
    \texttt{<s> \{sentence start\}} \texttt{\{candidate\} \{sentence end\} </s>} & \xmark \\ \midrule
    
    \multirow{2}{*}{\textbf{XCOPA}} &
    \texttt{<s> \{sentence\}} because \texttt{\{candidate answer\} </s>} \\ &
    \texttt{<s> \{sentence\}} so \texttt{\{candidate answer\} </s>}
    
    & \xmark \\ \midrule
    
    \multirow{2}{*}{\textbf{Hate Speech}} &
    
    \multirow{2}{*}{\texttt{<s> The sentence is \{label\}}. \texttt{\{sentence\} </s>} } &
    
    sexist, racist, offensive, abusive, hateful (Positive) \\ & & normal, common, ok, usual, acceptable (Negative)

    \\ \midrule
    \multirow{2}{*}{\textbf{NER}} & 
    \multirow{2}{*}{\texttt{<s>}lang: \texttt{\{lang\}} \textbackslash n Tagged sentence: \texttt{\{sentence with tags\}}} &

    I-LOC, I-MISC, \\ & & I-ORG, I-PER, O 
    \\ \midrule
    
    \multirow{4}{*}{\textbf{POS}} & 
    \multirow{4}{*}{\texttt{<s>}lang: \texttt{\{lang\}} \textbackslash n Tagged sentence: \texttt{\{sentence with tags\}}}  &

    ADJ, ADP, ADV, AUX, \\ & & CCONJ, DET, INTJ, NOUN, \\ & & NUM, PART, PRON, PROPN, PUNCT, \\ & & SCONJ, SYM, VERB, X
    \\ \bottomrule
    \end{tabular}
    }
    \caption{Prompt examples for each downstream task. The examples are in English for illustration purposes.}
    \label{tab:prompts}
\end{table*}
    \begin{table}[t!]
\scriptsize
\centering
\setlength{\tabcolsep}{1pt}
\begin{tabular}{lcccccc}
\toprule
\textbf{Model} & \textbf{$k$-shot} & \textbf{XWINO} & \textbf{PAWSX} & \textbf{XCOPA} & \textbf{XNLI} & \textbf{Hate Speech} \\ \midrule
\multirow{4}{*}{mGPT$_\text{1.3B}$} & $0$    &   56.2 & \underline{53.1} & 55.5 & 40.6 & 50.0    \\
         & $1$    &   57.0 & 51.3 & 54.9 & 36.1 &   \xmark      \\ 
         & $4$    &   56.8 & 52.2 & 54.8 & 37.4 & 50.8     \\
         & $16$   & 54.5 & 52.2 & 54.8 & 37.9 & \xmark     \\ \midrule
\multirow{4}{*}{mGPT$_\text{13B}$}  & $0$    &   59.3 & 51.5 & 58.2 & \underline{42.6} & \textbf{53.1}    \\
         & $1$    &   61.0 & 50.6 & 57.9 & 37.5  &   \xmark      \\
         & $4$    &   61.8 & 51.6 & 58.3 & 41.4 & 51.5     \\
         & $16$   &   59.2 & \textbf{55.1} & 57.3 & 33.3 &    \xmark     \\ \midrule
\multirow{4}{*}{XGLM$_\text{1.7B}$} & $0$    &   54.2  &   50.3  &   55.5  &   \underline{42.6} &  50.1     \\
         & $1$    &   58.0  &    45.9  &   56.8  & 36.4   & \xmark     \\
         & $4$    &   57.9  &    45.9  &   56.2  & 38.8   &  49.5     \\
         & $16$   &   \xmark & 44.2 &   56.1  &   36.5 &   \xmark      \\ \midrule
\multirow{4}{*}{XGLM$_\text{7.5B}$} & $0$    &   59.2  &   50.1  &   55.5  &   \textbf{44.7} &  50.1     \\
         & $1$    & \underline{63.7} &   46.4  &   60.6  & 36.9   &  \xmark    \\
         & $4$    &   \textbf{64.2}  &   45.3  &   \underline{61.4}  &   40.1 &  \underline{51.8}     \\
         & $16$   & \xmark   &   44.9  &   \textbf{62.5}  & 40.0   &    \xmark     \\ \bottomrule
\end{tabular}

\caption{Accuracy scores (\%) on classification tasks averaged across languages.}
\label{tab:classification}
\end{table}

\paragraph{Results by Language Family} Analyzing results by the language family (see~\autoref{fig:family_wise_ppl}), we find that mGPT$_\text{13B}$ shows consistently lower perplexities as opposed to mGPT$_\text{1.3B}$. Specifically, mGPT$_\text{1.3B}$ underperforms mGPT$_\text{13B}$ on Basque, Greek, Kartvelian, and Turkic families.

\paragraph{Correlation Analysis} We present the results in~\autoref{table:corr_res}. We observe that the language modeling performance depends on the language script and model size. In particular, the non-Latin languages receive lower scores on average, while mGPT$_\text{13B}$ performs better than mGPT$_\text{1.3B}$ in this setting. However, the positive correlation between the pretraining corpus size and perplexity in particular languages can be attributed to the low diversity of the text domains in the pretraining monolingual corpora for the low-resource languages. Such corpora contain Wikipedia articles on a limited amount of general topics; therefore, the model learns the distribution in the corpora without being able to generalize well. In general, the results align with~\citet{workshop2023bloom}, who report that the considered criteria can affect the knowledge acquired by BLOOM$_\text{1B}$ and BLOOM$_\text{176B}$.

\subsection{Downstream Evaluation} 
\label{subsection:downstream_evaluation}
We conduct an extrinsic evaluation of mGPT and baselines on classification and sequence labeling tasks in zero-shot and few-shot settings. In the zero-shot setting, the model is shown a test example formatted as a prompt in natural language, while in the few-shot setting, the model is provided with $k$ demonstrations from the training data specified via prompts. The prompt examples for each task are presented in~\autoref{tab:prompts}.

\subsubsection{Classification}
\label{subsec:classification}
\paragraph{Tasks} The classification tasks include commonsense reasoning (XCOPA;~\citealp{ponti-etal-2020-xcopa}), natural language inference (XNLI;~\citealp{conneau-etal-2018-xnli}), Winograd schema challenge (XWINO;~\citealp{tikhonov-ryabinin-2021-heads}), paraphrase detection (PAWSX;~\citealp{yang-etal-2019-paws}), and hate speech detection~\cite{davidson2017automated}.

\paragraph{Method} mGPT utilizes per-token cross-entropy loss, which is reduced to negative log probability due to one-hot encoding of the tokens. We select the target label associated with the prompt that results in the lowest sum of negative log probabilities for its tokens. The few-shot experiments are run five times with different random seeds, while the zero-shot experiments are run only once since the model loss is determined.

\paragraph{Baselines} The XGLM$_\text{1.7B}$ and XGLM$_\text{7.5B}$ models are used as the baselines in the classification experiments. We reproduce the XGLM evaluation based on the methodology by~\citet{lin2022fewshot} and use the model weights and code available in the \texttt{fairseq}\footnote{\href{https://github.com/pytorch/fairseq/tree/main/examples/xglm}{\texttt{github.com/pytorch/fairseq/xglm}}} library~\cite{ott-etal-2019-fairseq}. We select prompts according to the templates reported by~\citeauthor{lin2022fewshot}. Prompts for non-English languages are automatically translated with Google Translate. 


\paragraph{Results} \autoref{tab:classification} presents the classification results averaged across languages. The ``\xmark'' tag marks $k$-shot settings not reported by~\citeauthor{lin2022fewshot}. We do not perform them for reproducibility purposes and fair comparison. The results by~\citeauthor{lin2022fewshot} are reproduced in the zero-shot setup, and some scores are even slightly higher. However, not all results are reproduced, e.g., PAWSX and XNLI. We attribute this to potential differences in the translated prompts.

Overall, we observe that mGPT$_\text{1.3B}$ is comparable with XGLM$_\text{1.7B}$ while having fewer weights and is pretrained in twice as many languages. mGPT$_\text{13B}$ performs better than  XGLM$_\text{7.5B}$ in zero-shot setting on all tasks except XNLI. At the same time, it lags behind in a few-shot setting being better than XGLM$_\text{7.5B}$ only in XNLI and PAWSX tasks. Comparing the performance across languages, we find that English receives the highest accuracy for all tasks. The mGPT$_\text{1.3B}$ and mGPT$_\text{13B}$ models show high accuracy for the Austronesian, Dravidian, Japonic, Germanic, and Romance language families. Only the Afro-Asiatic family gets low accuracy. The mGPT models perform better than the XGLM counterparts for Austronesian, Koreanic, and Romance languages.

Our results on hate speech detection are consistent with~\citeauthor{lin2022fewshot}. The performance is slightly better across the five languages but still close to random guessing (see~\autoref{tab:classification_hatespeech}). The manual analysis shows that the behavior is sensitive to the input prompts, most notably for Polish. Increasing the number of demonstrations can lead to performance degradation on some classification tasks for both mGPT and XGLM.

\begin{table}[t!]
\centering
\scriptsize
\setlength{\tabcolsep}{4pt}
\begin{tabular}{lccccccc}
\toprule
\textbf{Model} & \textbf{$k$-shot} & \textbf{en} & \textbf{es} & \textbf{pt} & \textbf{pl} & \textbf{it} \\ \midrule
\multirow{2}{*}{mGPT$_\text{1.3B}$} & $0$ & 55.1 & 52.1 & 42.3 & 50.0 & 50.2 \\ 
  & $4$ & 50.1 & 50.2 & 51.7 & \underline{51.5} & 50.4 \\ 
 \midrule
\multirow{2}{*}{mGPT$_\text{13B}$} & $0$ & \underline{59.0} & \textbf{55.2} & 46.9 & 50.0 & \textbf{54.6} \\ 
 & $4$ & 52.2 & 50.0 & 50.8 & \textbf{53.4} & 51.0 \\
\midrule
\multirow{2}{*}{XGLM$_\text{1.7B}$} & $0$ & 54.8 & 51.8 & \underline{52.3} & 50.0 & \underline{54.5} \\ 
 & $4$ &  51.0 & 48.8 & 49.2 & 46.7 & 51.0 \\ \midrule
\multirow{2}{*}{XGLM$_\text{7.5B}$} & $0$ & \textbf{61.7} & \underline{52.4} & \textbf{52.3} & 50.0 & 49.0 \\ 
 & $4$ & 51.8 & 51.3 & 51.5 & 51.4 & 52.9 \\ \bottomrule
\end{tabular}
\caption{Accuracy scores (\%) on hate speech detection by language. The best score is put in bold, the second best is underlined.}
\label{tab:classification_hatespeech} 
\end{table}

\subsubsection{Sequence Labeling}
\label{subsection:sequence_labeling}
\paragraph{Tasks} The sequence labeling tasks include named entity recognition (NER) and part-of-speech tagging (POS) from the XGLUE benchmark~\cite{liang-etal-2020-xglue}. To address other medium-resource and resource-lean languages, we use the Universal Dependencies treebanks (UD;~\citealp{nivre-etal-2016-universal}) to evaluate POS-tagging in Armenian, Belarusian, Buryat, Kazakh, Tatar, Ukrainian, and Yakut.

\begin{table}[th!]
\centering
\scriptsize
\begin{tabular}{lcccccc}
\toprule 
\textbf{Model}  &  \textbf{de} &  \textbf{en}&  \textbf{es}&  \textbf{nl}  & \textbf{Avg.}\\ \midrule
Random  & 1.9 & 3.1 & 1.8 & 1.6 & 2.1\\ \midrule
mGPT$_\text{1.3B}$ & 12.2 & 22.1 & 12.7 & 13.1 & 15.0 \\ 
      
mGPT$_\text{13B}$ & 5.6 & 20.9 & 10.4 & 6.7 & 10.9\\ \midrule

M-BERT$_\text{base}$  & 69.2 & 90.6 & \underline{75.4} & 77.9 & 78.2\\ 

XLM-R$_\text{base}$  & 70.4 & \underline{90.9} & \textbf{75.2} & \underline{79.5} & \underline{79.0}\\ 

Unicoder  & \textbf{71.8} & \textbf{91.1} & 74.4 & \textbf{81.6} & \textbf{79.7}

\\ \bottomrule
\end{tabular}
\caption{F1-scores for NER by language. The mGPT models are evaluated in the 4-shot setting. The best score is put in bold, the second best is underlined.}
\label{tab:ner-results-fair} 
\end{table}

\paragraph{Method} We use a modified approach to the sequence labeling tasks compared to \S\ref{subsec:classification}. 
Given a sentence of $n$ words, we iteratively predict the label for each word $x_i$ using the preceding words $x_{<i}$ and their predicted labels $l_{<i}$ as the context using a template ``$x_{<i}l_{<i}$\_'', where $i$ is the current token index and ``\_'' is a placeholder. The only exception is the first token $x_i$ used as the context. The placeholder is filled with each possible target label $l \in L$ at each step. We select the label with the lowest sum of losses per token in the resulting string. The experiments are run in the zero-shot and $4$-shot settings\footnote{We report the results only in the $4$-shot setting since the manual analysis reveals that the models have failed to capture the task, giving constant predictions without any additional examples.}.

\paragraph{Example} Consider an example for the POS-tagging task \textsc{``I [PRON] want [VERB] it [PART] . [PUNCT]''}, which requires $4$ procedure steps. First, we combine the placeholder in the string \textsc{``I\_''} with each possible POS tag and select the most probable candidate. Next, we repeat the procedure for \textsc{``I\_$l_i$ want\_''} and so on.

\begin{table*}[h!]
\centering
\resizebox{1\linewidth}{!}{
\tiny
\setlength{\tabcolsep}{1.5pt}
\begin{tabular}{lccccccccccccccccccc|ccccccc}
\toprule 
\multirowcell{2}[-0.5ex]{\textbf{Model}}  &
\multicolumn{18}{c}{\textbf{XGLUE}} & \multicolumn{7}{c}{\textbf{CIS \& Low-Resource UD}} \\
\cmidrule{2-27}

& \textbf{ar} &  \textbf{bg} &  \textbf{de}&  \textbf{el} &  \textbf{en} &  \textbf{es} &\textbf{fr} &  \textbf{hi}&  \textbf{it}&  \textbf{nl} &  \textbf{pl} &  \textbf{pt} &  \textbf{ru}&  \textbf{th}&  \textbf{tr} &  \textbf{ur}&  \textbf{vi} &\textbf{zh} & \textbf{Avg.} & \textbf{be} &  \textbf{bxr}&  \textbf{hy}&  \textbf{kk}  &  \textbf{sah} &  \textbf{tt}&  \textbf{uk}  \\ \midrule

Random   & 6.5 &  6.5 &  6.0 &  5.2 &  4.4 &  5.7 & 5.5 &  6.7 & 6.6 &  6.6 &  5.9 & 4.7 & 6.0 &  6.4 &  6.8 &  1.2 &  7.0 &  7.1 & 5.8 & 1.3 & 5.7 & 5.9  & 2.6 & 9.6 & 8.7 & 4.8  \\ \midrule
mGPT$_\text{1.3B}$& 16.5 &  24.5 &  30.6 & 20.9 &  40.0 &  24.3 & 27.0 &  16.2 &  25.4 &  28.8 &  28.3 & 24.6 &  29.4 &  12.9 &  30.4 &  15.0 &  25.6 & 19.5 & 24.4 & \textbf{21.5} & \textbf{28.4} & \textbf{14.7}  & \textbf{22.8} & \textbf{19.9} & \textbf{21.4} & \textbf{22.5} \\ 
      
mGPT$_\text{13B}$  & 11.7 &  21.8 &  26.8 &  16.1 & 36.0 &  22.2 & 25.0 &  12.3 &  26.5 &  26.5 &  24.2 & 21.8 & 21.8 &  9.5 &  26.8 &  12.7 &  21.5 &  12.5 & 20.9 & \underline{10.6} & \underline{7.7} & \underline{7.3}  & \underline{9.4} & \underline{11.8} & \underline{9.2} & \underline{10.9}  \\ \midrule

M-BERT$_\text{base}$  & 52.4 & 85.0 & 88.7 & 81.5 & 95.6 & 86.8 & 87.6 & 58.4 & 91.3 & 88.0 & 81.8 & 88.3 & 78.8 & 43.3 & 69.2 & 53.8 & 54.3 & 58.3 & \underline{74.7} & \xmark & \xmark & \xmark  & \xmark & \xmark & \xmark & \xmark
\\ 

XLM-R$_\text{base}$  & \underline{67.3} & \textbf{88.8} & \textbf{92.2} & \textbf{88.2} & \textbf{96.2} & \textbf{89.0} & \textbf{89.9} & \textbf{74.5} & \textbf{92.6} & \textbf{88.5} & \textbf{85.4} & \textbf{89.7} & \textbf{86.9} & \textbf{57.9} & \underline{72.7} & \textbf{62.1} & \underline{55.2} & \textbf{60.4} & \textbf{79.8} & \xmark & \xmark & \xmark  & \xmark & \xmark & \xmark & \xmark \\ 


Unicoder & \textbf{68.6} & \underline{88.5} & \underline{92.0} & \underline{88.3} & \underline{96.1} & \underline{89.1} & \underline{89.4} & \underline{69.9} & \underline{92.5} & \underline{88.9} & \underline{83.6} & \underline{89.8} & \underline{86.7} & \underline{57.6} & \textbf{75.0} & \underline{59.8} &\textbf{56.3} & \underline{60.2} & \textbf{79.6} & \xmark & \xmark & \xmark  & \xmark & \xmark & \xmark & \xmark 

\\ \bottomrule
\end{tabular}
}
\caption{Accuracy scores (\%) for XGLUE and Universal Dependencies POS-tagging by language. mGPT models are evaluated in the 4-shot setting. The best score is put in bold, the second best is underlined.}
\label{tab:pos-results-detailed} 
\end{table*}

\begin{figure*}[!ht]
    \centering
    \includegraphics[width=1.0\textwidth]{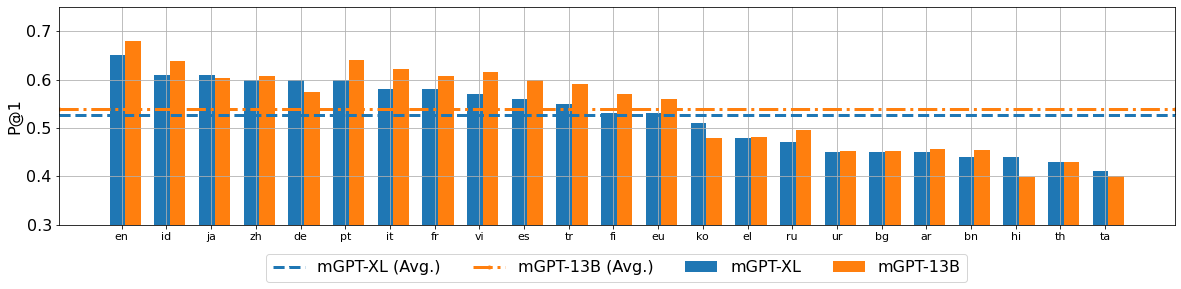}
    \caption{Knowledge probing results for 23 languages. The performance of a random baseline is 0.33.}
    \label{fig:mlama_probing}
\end{figure*}

\paragraph{Baselines} We use results reported in~\citeauthor{liang-etal-2020-xglue} as the baselines: M-BERT, XLM-R, and Unicoder~\cite{huang-etal-2019-unicoder}. Note that the baselines are \emph{finetuned} on the corresponding training set. The performance is evaluated with the F1-score (NER) and the accuracy score (POS-tagging)\footnote{We evaluate the sequence labeling tasks using the XGLUE code: \href{https://github.com/microsoft/XGLUE/tree/master/evaluation}{\texttt{github.com/microsoft/XGLUE.}}} according to the XGLUE methodology.

\paragraph{NER Results} ~\autoref{tab:ner-results-fair} shows counterintuitively that mGPT$_\text{1.3B}$ outperforms mGPT$_\text{13B}$ on all languages. 4-shot falls behind finetuned models but significantly outperforms random guessing for both mGPT models. Per-language language analysis shows a large gap between English and other languages (for mGPT$_\text{13B}$ the F1-score on English is more than twice higher than for any of the other languages), while for German, both models perform the worst. This pattern coincides with the baseline results. In addition, it could be noted that while for mGPT$_\text{1.3B}$ the F1-score exceeds the 10 percent threshold for all languages, this is not the case for mGPT$_\text{13B}$.

\paragraph{POS-tagging Results} POS-tagging results for XGLUE benchmark and resource-lean languages are presented in~~\autoref{tab:pos-results-detailed}. Similarly to the NER task, mGPT$_\text{1.3B}$ outperforms mGPT$_\text{13B}$ practically in all languages except for Italian. On average mGPT$_\text{1.3B}$ achieves accuracy score of 0.24 while mGPT$_\text{13B}$ only scores 0.21. These results are still far behind fine-tuned models; however, they are significantly higher than random guessing. Analyzing the results for the low-resource languages, it can be seen that mGPT$_\text{1.3B}$ performance is comparable with its performance on XGLUE, while the mGPT$_\text{13B}$ scores are lower.

\subsection{Knowledge Probing} 
\label{subsection:probing}
\paragraph{Method} We probe our models for factual knowledge in 23 languages using the mLAMA dataset~\cite{kassner-etal-2021-multilingual}. The task is to complete a knowledge triplet \textit{<subject, relation, object>} converted to templates for querying LMs. Consider an example from the original LAMA~\cite{petroni-etal-2019-language} for English, where \textit{<Dante, born-in, X>} is converted to the template \textit{``Dante was born in [MASK]''}. We follow~\citeauthor{lin2022fewshot} to design the probing task. As each such query contains hundreds of negative candidates on average, we limit the number of candidates to three, i.e., one is the ground truth candidate and the other two candidates are randomly sampled from the provided knowledge source. The probing performance is evaluated with precision@1 averaged over all relations per language.

\begin{figure*}[!ht]
    \centering
    \includegraphics[width=.97\textwidth]
    {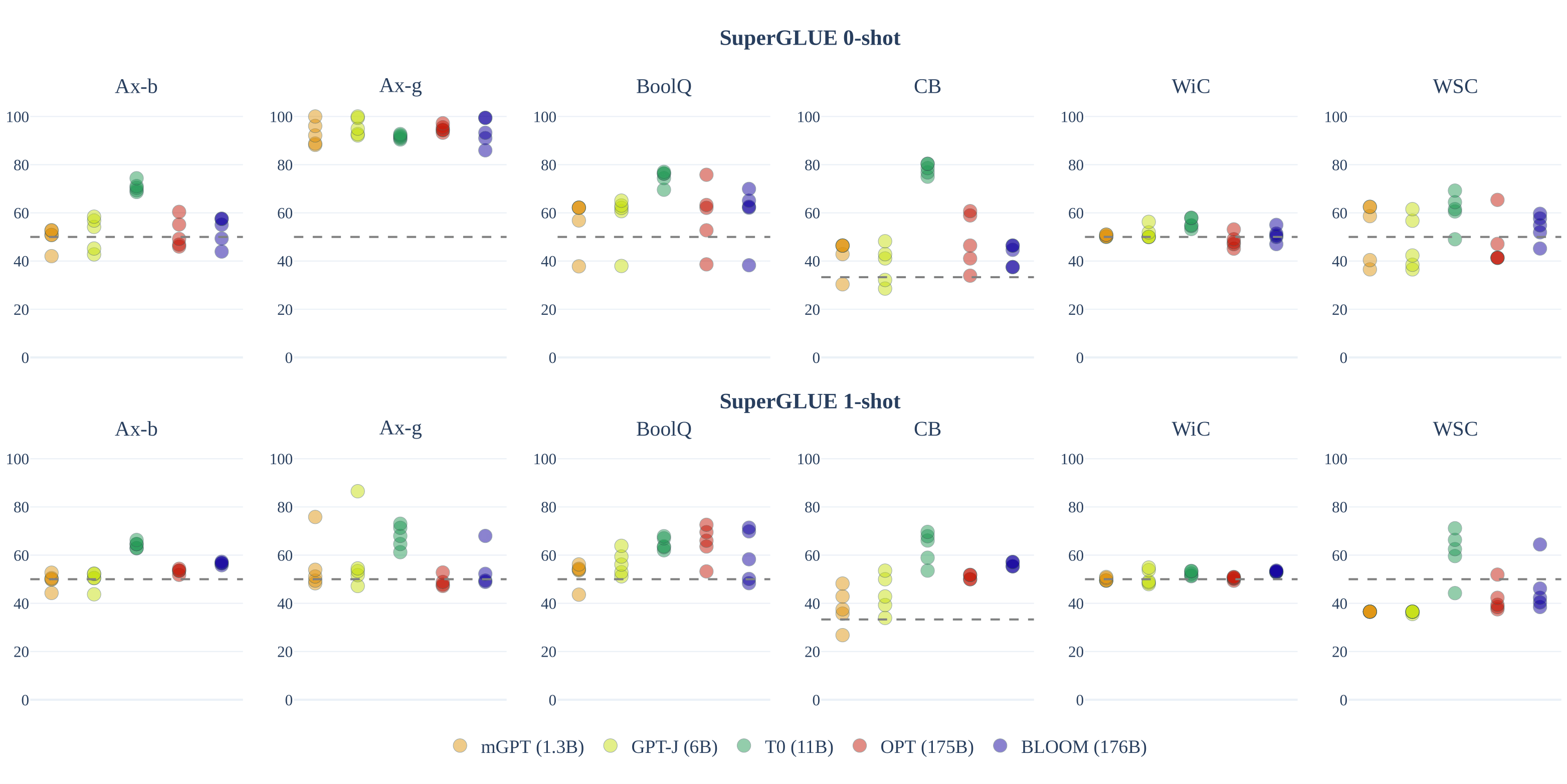}
    \caption{The SuperGLUE evaluation results in the zero-shot and one-shot settings~\cite{workshop2023bloom}.}
    \label{fig:superglue}
\end{figure*}

\begin{table*}[ht!]
\resizebox{1\linewidth}{!}{
\tiny
\centering
\begin{tabular}{ccccccccc}
\toprule
\textbf{ISO} & \textbf{Avg.  length}   & \textbf{Distinct}$_\text{1}$ & \textbf{Vocabulary size} & \textbf{Unique}$_\text{1}$  & \textbf{Entropy}$_\text{1}$ & \textbf{TTR} & \textbf{MSTTR} \\ \midrule
en & \cellcolor[HTML]{FFFFFF}39.13 \tiny{$\pm$ 22.61}   & 0.071   & 387  & 103   & 6.175  & 0.097    & 0.228  \\
fr    &  \cellcolor[HTML]{FFFFFF}23.53 \tiny{$\pm$ 17.92} & 0.128   & 486  & 181   & 6.875  & 0.159    & 0.346  \\
de    &  \cellcolor[HTML]{FFFFFF}30.85 \tiny{$\pm$ 17.33} & 0.113   & 453  & 159   & 6.850  & 0.151    & 0.340  \\
es    &  \cellcolor[HTML]{FFFFFF}12.71 \tiny{$\pm$ 15.54} & 0.102   & 413  & 124   & 6.818  & 0.148    & 0.315  \\
zh    & 3.157 \tiny{$\pm$ 2.39}  & 0.492   & 188  & 124   & 7.055  & 0.525    & 0.526    \\ \bottomrule  
\end{tabular}
}
\caption{The results for lexical diversity of generated texts on the GEM story generation task.}
\label{table:stats2}
\end{table*}

\paragraph{Results} \autoref{fig:mlama_probing} outlines the results for mGPT$_\text{1.3B}$ and mGPT$_\text{13B}$. The overall pattern is that the performance is equal to or above 0.6 for Germanic, Romance, Austro-Asiatic, Japonic, and Chinese languages. However, Uralic, Slavic, Koreanic, and Afro-Asiatic languages receive scores of lower than 0.5. We also find that scaling the number of model parameters usually boosts the performance for high-resource languages up to 5 points, while no significant improvements are observed in the other languages. Comparing our results with~\citeauthor{lin2022fewshot}, we conclude that our models achieve lower performance than XGLM$_\text{7.5B}$ almost in all languages and perform on par with GPT3-Curie$_\text{6.5B}$.

\subsection{External Evaluation}
\label{results:bloom}
\paragraph{General Language Understanding} \citet{workshop2023bloom} compared the performance of BLOOM$_\text{176B}$, mGPT$_\text{1.3B}$, OPT$_\text{175B}$~\cite{zhang2022opt}, GPT-J$_\text{6B}$~\cite{wang2021gpt}, and T0$_\text{11B}$~\cite{victor2022multitask} on subset of tasks from the SuperGLUE benchmark~\cite{wang2019superglue} in the zero-shot and one-shot settings. The results of evaluating the models using five prompts are presented in~\autoref{fig:superglue}. The mGPT$_\text{1.3B}$ model has comparable performance despite having fewer weights. In the zero-shot setting, the performance of mGPT$_\text{1.3B}$, BLOOM$_\text{176B}$, OPT$_\text{175B}$, and GPT-J$_\text{6B}$ on the considered tasks is above random guessing. We also observe the strong performance of mGPT$_\text{1.3B}$ on the Winogender Schema Diagnostics (Ax-g). 
In the one-shot setting, mGPT$_\text{1.3B}$ performs on par with GPT-J$_\text{6B}$, and the resulting variability is significantly reduced across all prompts. 

\paragraph{Multilingual Clause-level Morphology} The first shared task on Multilingual Clause-level Morphology~\cite{goldman-etal-2022-mrl} covers nine languages and includes three sub-tasks: (i) inflection (generating a word form given a lexeme and a set of morphosyntactic features), (ii) reinflection (reinflect an input sentence according to a given set of morphosyntactic features), and (iii) detect a root and its features in an input sentence. ~\citet{acikgoz-etal-2022-transformers} develop a first-place solution based on mGPT$_\text{1.3B}$ and prefix-tuning method, outperforming other solutions and baselines on the third task. 

\subsection{Generation Evaluation}

\paragraph{Method} We compute seven lexical diversity metrics from~\citet{gehrmann-etal-2021-gem} using the mGPT outputs\footnote{We use the generation hyperparameters: $temperature=1$, $max\_length=100$, $top\_k=5$, $top\_p=0.9$.} on $100$ test set samples from the story generation task in five languages: English, French, German, Spanish, and Chinese~\cite{chen-etal-2022-mtg}. The diversity metrics include the Shannon Entropy over unigrams (Entropy$_\text{1}$), the mean segmented type-token ratio over segment lengths of 100 (MSTTR), the ratio of distinct unigrams over the total number of unigrams (Distinct$_\text{1}$), and the counter of unigrams that appear once in the collection of generated outputs (Unique$_\text{1}$).

\paragraph{Results} The results are presented in~\autoref{table:stats2}. The diversity metrics scores for Chinese are the highest, while the mean generated text length is the shortest. This is likely due to its logographic writing. The results for the Indo-European languages are similar (French, German, and Spanish), indicating that mGPT$_\text{1.3B}$ generates diverse texts in these languages. Surprisingly, the metrics are lower for English, with the average text length being longer. Our current natural language generation evaluation approach lacks downstream tasks, which we leave for future work. 


\section{Discussion}
Our key takeaways on pretraining and evaluating large-scale multilingual autoregressive LMs are summarized below.

\subsection{Model Scaling}
\paragraph{Empirical Results} The language modeling results for mGPT$_\text{1.3B}$ and mGPT$_\text{13B}$ suggest that the model scaling improves its generation abilities for all given languages (see~\S\ref{subsection:language_modeling}). However, it does not improve performance on the downstream and probing tasks  (see~\S\ref{subsection:downstream_evaluation};~\S\ref{subsection:probing}). Overall, the language modeling performance depends on the model size and the pretraining corpus size in a language, and smaller models may better encode linguistic information than larger ones. These findings align with \citet{workshop2023bloom}. 

\paragraph{Takeaways} Our work had been conducted a year before the Chinchilla scaling laws were introduced~\cite{hoffmann2022training}. According to the advanced methods of scaling LMs, our pretraining corpus can be sufficiently extended to improve the generalization abilities of the mGPT$_\text{13B}$ model. At the same time, the pretraining corpus design can promote the model underfitting and overfitting on particular languages. We believe it can be accounted for by aggregating the language-specific cross-entropy loss and producing language weights similar to~\citet{xie2023doremi}.

\subsection{Lack of Data}
\paragraph{Empirical Results} Another challenging factor is the lack of high-quality data for the low-resource languages. Although mGPT shows promising results on the language modeling and sequence labeling tasks for the underrepresented languages (see~\S\ref{subsection:language_modeling},~\S\ref{subsection:downstream_evaluation}), the low amount of evaluation resources limits the scope of analyzing the model generalization abilities. The correlation between the model performance and the amount of pretraining data in a language (see~\S\ref{subsection:language_modeling}, and,~\citealp[e.g.,][]{lauscher-etal-2020-zero,ahuja-etal-2022-multi}) further highlights the need for creating text corpora in such languages.

\paragraph{Takeaways} The question of addressing the discrepancy in data distribution across the world's languages remains unresolved. Our data collection and filtration approach is equivalent for all considered languages. Extending the language-agnostic heuristics is restrained due to the lack of linguistic expertise. However, we assume that experimenting with the training data for the text quality classifiers can improve the resulting quality of the corpora for the low-resource languages (e.g., training the classifiers on different mixtures of data in the medium and high-resource languages). 

\begin{table}[t!]
\centering
\tiny
\begin{tabular}{lll}
\toprule
\textbf{Language}  & \textbf{HuggingFace URL} & \textbf{PPL} \\
\midrule
Armenian & \href{https://huggingface.co/ai-forever/mGPT-1.3B-armenian}{\texttt{hf.co/ai-forever/mGPT-1.3B-armenian}} & $1.7$\\
Azerbaijan &  \href{https://huggingface.co/ai-forever/mGPT-1.3B-azerbaijan}{\texttt{hf.co/ai-forever/mGPT-1.3B-azerbaijan}} & $5.4$ \\
Bashkir & \href{https://huggingface.co/ai-forever/mGPT-1.3B-bashkir}{\texttt{hf.co/ai-forever/mGPT-1.3B-bashkir}} & $7.1$ \\
Belorussian & \href{https://huggingface.co/ai-forever/mGPT-1.3B-belorussian}{\texttt{hf.co/ai-forever/mGPT-1.3B-belorussian}} & $27.7$ \\
Bulgarian  & \href{https://huggingface.co/ai-forever/mGPT-1.3B-belorussian}{\texttt{hf.co/ai-forever/mGPT-1.3B-belorussian}} & $15.2$ \\
Buryat & \href{https://huggingface.co/ai-forever/mGPT-1.3B-buryat}{\texttt{hf.co/ai-forever/mGPT-1.3B-buryat}} & $17.6$  \\
Chuvash & \href{https://huggingface.co/ai-forever/mGPT-1.3B-chuvash}{\texttt{hf.co/ai-forever/mGPT-1.3B-chuvash}} & $28.8$  \\ 
Georgian & \href{https://huggingface.co/ai-forever/mGPT-1.3B-georgian}{\texttt{hf.co/ai-forever/mGPT-1.3B-georgian}} & $16.9$ \\ 
Kalmyk & \href{https://huggingface.co/ai-forever/mGPT-1.3B-kalmyk}{\texttt{hf.co/ai-forever/mGPT-1.3B-kalmyk}} & $14.0$ \\
Kazakh & \href{https://huggingface.co/ai-forever/mGPT-1.3B-kazakh}{\texttt{hf.co/ai-forever/mGPT-1.3B-kazakh}} & $3.4$  \\ 
Kirgiz & \href{https://huggingface.co/ai-forever/mGPT-1.3B-kirgiz}{\texttt{hf.co/ai-forever/mGPT-1.3B-kirgiz}} & $8.2$ \\ 
Mari & \href{https://huggingface.co/ai-forever/mGPT-1.3B-mari}{\texttt{hf.co/ai-forever/mGPT-1.3B-mari}} & $21.2$ \\ 
Mongol & \href{https://huggingface.co/ai-forever/mGPT-1.3B-mongol}{\texttt{hf.co/ai-forever/mGPT-1.3B-mongol}} & $4.4$ \\
Ossetian  &  \href{https://huggingface.co/ai-forever/mGPT-1.3B-ossetian}{\texttt{hf.co/ai-forever/mGPT-1.3B-ossetian}} & $18.7$ \\ 
Persian & \href{https://huggingface.co/ai-forever/mGPT-1.3B-persian}{\texttt{hf.co/ai-forever/mGPT-1.3B-persian}} & $33.4$\\ 
Romanian & \href{https://huggingface.co/ai-forever/mGPT-1.3B-romanian}{\texttt{hf.co/ai-forever/mGPT-1.3B-romanian}} & $3.4$ \\ 
Tajik & \href{https://huggingface.co/ai-forever/mGPT-1.3B-tajik}{\texttt{hf.co/ai-forever/mGPT-1.3B-tajik}} & $6.5$ \\ 
Tatar & \href{https://huggingface.co/ai-forever/mGPT-1.3B-tatar}{\texttt{hf.co/ai-forever/mGPT-1.3B-tatar}} & $3.7$ \\ 
Turkmen & \href{https://huggingface.co/ai-forever/mGPT-1.3B-turkmen}{\texttt{hf.co/ai-forever/mGPT-1.3B-turkmen}} & $28.5$ \\ 
Tuvan & \href{https://huggingface.co/ai-forever/mGPT-1.3B-tuvan}{\texttt{hf.co/ai-forever/mGPT-1.3B-tuvan}} & $40.8$ \\ 
Ukranian & \href{https://huggingface.co/ai-forever/mGPT-1.3B-ukranian}{\texttt{hf.co/ai-forever/mGPT-1.3B-ukranian}} & $7.1$  \\ 
Uzbek & \href{https://huggingface.co/ai-forever/mGPT-1.3B-uzbek}{\texttt{hf.co/ai-forever/mGPT-1.3B-uzbek}} & $6.8$ \\ 
Yakut & \href{https://huggingface.co/ai-forever/mGPT-1.3B-yakut}{\texttt{hf.co/ai-forever/mGPT-1.3B-yakut}} & $10.6$ \\

\bottomrule
\end{tabular}
\caption{A list of the mGPT$_\text{1.3B}$ models continuously pretrained on monolingual corpora for $23$ languages.}
\label{tab:monolingual_ft}
\end{table}

As the follow-up work, we release $23$ versions of the mGPT$_\text{1.3B}$ model continuously pretrained with language modeling objective on monolingual corpora for medium-resource and low-resource languages collected through collaboration with the NLP community.~\autoref{tab:monolingual_ft} summarizes the models by language and the language modeling performance on the held-out monolingual test sets. Examples of the corpora include Eastern Armenian National Corpus~\cite{khurshudyan-etal-2022-eastern}, OpenSubtitles~\cite{lison-tiedemann-2016-opensubtitles2016}, and TED talks. Continued pretraining on additional data improves the language modeling performance.


\subsection{Language Selection}
\paragraph{Empirical Results} Results of mGPT$_\text{1.3B}$ on most of the classification tasks are on par or better than the results of the XGLM$_\text{1.7B}$ given that mGPT covers twice as many languages (see~\S\ref{subsection:downstream_evaluation}). However, mGPT underperforms the baselines on several multi-class classification and probing tasks. 

\paragraph{Takeaways} We find that balancing the pretraining corpus by the language family helps improve the language modeling abilities for underrepresented languages due to their typological similarity with the medium and high-resource languages (see~\S\ref{subsection:language_modeling}). However, increasing language diversity can lead to performance degradation because of the curse of multilinguality and a limited model capacity~\cite{conneau-etal-2020-unsupervised}. 

\subsection{Tokenization} 
\paragraph{Empirical results} We conduct an ablation study to analyze the impact of the tokenization strategy on language modeling performance. We find that the considered strategies do not improve the model's perplexity. However, the main drawback of the perplexity-based evaluation is that it only partially assesses the model generalization abilities.

\paragraph{Takeaways} The optimal tokenization method and vocabulary size remain an open question, particularly in the multilingual setup~\cite{mielke2021words}. There are no established methods for defining the vocabulary size based on the amount of textual data in different languages. Our experiments are limited to a fixed vocabulary size, and we leave further investigation of the tokenization strategies and their configurations for future work.

\subsection{Zero-shot and Few-shot Performance}
\paragraph{Empirical results} 
\begin{itemize}[itemsep=-2pt,partopsep=1ex,parsep=1ex,leftmargin=1.5em]
    \item Increasing the number of demonstrations does not always lead to improvements but decreases the performance on some downstream tasks (see~\S\ref{subsec:classification}; \S\ref{subsection:sequence_labeling}). This observation aligns with~\citet{lin2022fewshot} and~\citet{NEURIPS2020_1457c0d6}.
    \item The zero-shot and few-shot performance may not exceed the random guessing on particular tasks, which points to the failure of a model to follow the guidance in the demonstration examples (see~\S\ref{subsec:classification}; \S\ref{subsection:sequence_labeling}).
    \item The prompting approach is unstable and hardly universal across languages, as indicated by the model sensitivity to the prompts.
    \item The mGPT models can assign higher probabilities to the most frequent tag in the input for the sequence labeling tasks (see~\S\ref{subsection:sequence_labeling}).
\end{itemize}

\paragraph{Takeaways} \begin{itemize}[itemsep=-2pt,partopsep=1ex,parsep=1ex,leftmargin=1.5em]
    \item The stability of the models with respect to the prompts may be improved using prompt-tuning~\cite{liu2023gpt} and contextual calibration~\cite{zhao2021calibrate} as shown in~\S\ref{results:bloom}.
    \item The generalization capabilities of the autoregressive LMs in sequence labeling tasks is an underexplored area. While our LMs achieve results higher than random guessing, the low performance can be attributed to the probability distribution shifts between the pretraining corpora and the prompts. We leave the investigation of the alternative prompt design~\cite{liu2023pre} and structured prediction methods~\cite{liu-etal-2022-autoregressive} for future work.
\end{itemize}

\section{Conclusion}
\label{sec:conclusion}
We introduce the mGPT$_\text{1.3B}$ and mGPT$_\text{13B}$ models, which cover $61$ languages from linguistically diverse $25$ language families. Our model is one of the first autoregressive LMs for economically endangered and underrepresented CIS and low-resource languages. The architecture design choices are based on the preliminary tokenization experiments and their perplexity-based evaluation. The model evaluation experiments include language modeling,  standardized cross-lingual NLU datasets and benchmarks, world knowledge probing, and social bias tasks. We evaluate the in-context learning abilities in zero and few-shot settings with a negative log-likelihood probability. We present a detailed analysis of the model performance, limitations, and ethical considerations. Despite the space for further quality growth and solving the highlighted limitations, the model shows significant potential and can become the basis for developing generative pipelines for languages other than English, especially the low-resource ones. This initiative has been developed for 23 diverse languages through collaboration with the NLP community. We hope to benefit cross-lingual knowledge transfer, annotation projection, and other potential applications for economically challenged and underrepresented languages and diversify the research field by shifting from the Anglo-centric paradigm.

\section{Ethical Statement and Social Impacts}
\subsection{Low-resource Languages} NLP for resource-lean scenarios is one of the leading research directions nowadays. The topic's relevance has led to proactive research on low-resource languages. Our work falls under this scope, introducing the first autoregressive LM for $61$ languages. To the best of our knowledge, we present one of the first attempts to address this problem for 20 languages of the Commonwealth of Independent States and the small peoples in Russia.

\subsection{Energy Efficiency and Usage} 
\label{ethical:co2}
Pretraining large-scale LMs requires many computational resources, which is energy-intensive and expensive. To address this issue, we used the sparse attention approach suggested by~\citet{NEURIPS2020_1457c0d6} and reduced the computational resources required to achieve the desired performance. The CO2 emission of pretraining the mGPT models is computed as Equation~\ref{eq:co2}~\cite{strubell-etal-2019-energy}:

\begin{equation}\label{eq:co2}
    CO2 = \frac{PUE * kWh * I^{CO2}}{1000}
\end{equation}

The power usage effectiveness ($PUE$) of our data centers is not more than $1.3$, the spent power is $30.6$k kWh (mGPT$_\text{1.3B}$) and $91.3$ kWh (mGPT$_\text{13B}$), and the CO2 energy intensity ($I^{CO2}$) in the region is $400$ grams per kWh. The resulting CO2 emission is $15.9$k kg (mGPT$_\text{1.3B}$) and $47.5$k kg (mGPT$_\text{13B}$). The emission is comparable with a single medium-range flight of a modern aircraft, which usually releases about $12$k kg of CO2 per $1$k km. Despite the costs, mGPT can be efficiently adapted to the user needs via few-shot learning, bringing down potential budget costs in the scope of applications in multiple languages, such as generating the content, augmenting labeled data, or summarizing news. The multilingual pretraining saves on data annotation and energy consumption, alleviating the carbon footprint. Model compression techniques, e.g., pruning and distillation, can reduce inference costs.

\subsection{Social Risks of Harm} Stereotypes and unjust discrimination present in pretraining corpora lead to representation biases in LMs. LMs can reflect historical prejudices against disadvantaged social groups and reproduce harmful stereotypes about gender, race, religion, or sexual orientation~\cite{weidinger2022taxonomy}. We have analyzed the mGPT's limitations on social risks of harm involving hate speech on the hate speech detection task. Our results are similar to~\citet{lin2022fewshot} in that the performance is close to random guessing. This may indicate a significant bias in the pretraining corpus, a mutual influence of languages during training, or methodological problems in the test set. We do not claim that our evaluation setup is exhaustive, and we assume that other biases can be revealed through a direct model application or an extended evaluation.

\subsection{Potential Misuse} The misuse potential of LMs increases with their ability to generate high-quality texts. Malicious users can perform a socially harmful activity that involves generating texts, e.g., spreading propaganda and other targeted manipulation~\cite{jawahar-etal-2020-automatic}. We recognize that our models can be misused in all supported languages. However, adversarial defense and artificial text detection models can mitigate ethical and social risks of harm. Our primary purpose is to propose multilingual GPT-style LMs for \textbf{research and development} needs, and we hope to work on the misuse problem with other developers and experts in mitigation research in the future.

\bibliography{anthology,tacl2021}
\bibliographystyle{acl_natbib}

\appendix
\newpage
\clearpage

\end{document}